\documentclass[runningheads]{llncs}

\usepackage[T1]{fontenc}
\def\doi#1{\href{https://doi.org/\detokenize{#1}}{\url{https://doi.org/\detokenize{#1}}}}
\usepackage{graphicx}
\usepackage{amsmath}
\usepackage{amssymb}
\usepackage{booktabs}
\usepackage{dsfont}
\usepackage[table,xcdraw]{xcolor}         
\usepackage{flushend}
\usepackage{appendix}
\usepackage{hyperref}
\usepackage{listings}
\DeclareMathOperator*{\argmax}{arg\,max}

\newcommand{\mb}[1]{{\boldsymbol{#1}}}

\newcommand{\trsp}{{\!\scriptscriptstyle\top}}

\newcommand\scalemath[2]{\scalebox{#1}{\mbox{\ensuremath{\displaystyle #2}}}}

\begin{document}
\title{RepsNet: Combining Vision with Language for Automated Medical Reports}
\titlerunning{RepsNet: Combining Vision with Language for Automated Medical Reports}
%
%
\author{Ajay K. Tanwani \and
Joelle Barral \and
Daniel Freedman }

\authorrunning{A. K. Tanwani et al.}
%
\institute{Verily \& Google Research, USA \\
\email{\{ajaytanwani,jbarral,danielfreedman\}@google.com}}
\maketitle              

\begin{abstract}
Writing reports by analyzing medical images is error-prone for inexperienced practitioners and time consuming for experienced ones. In this work, we present RepsNet that adapts pre-trained vision and language models to interpret medical images and generate automated reports in natural language. RepsNet consists of an encoder-decoder model: the encoder aligns the images with natural language descriptions via contrastive learning, while the decoder predicts answers by conditioning on encoded images and prior context of descriptions retrieved by nearest neighbour search. We formulate the problem in a visual question answering setting to handle both categorical and descriptive natural language answers. We perform experiments on two challenging tasks of medical visual question answering (VQA-Rad) and report generation (IU-Xray) on radiology image datasets. Results show that RepsNet outperforms state-of-the-art methods with $81.08 \%$ classification accuracy on VQA-Rad $2018$ and $0.58$ BLEU-1 score on IU-Xray. Supplementary details are available at: \url{https://sites.google.com/view/repsnet}

\end{abstract}

\section{Introduction}
\label{sec:intro}

\begin{figure}[t]
\begin{center}
\includegraphics[trim={0.0cm 0.0cm 0.0cm 0.1cm},clip,scale=0.415]{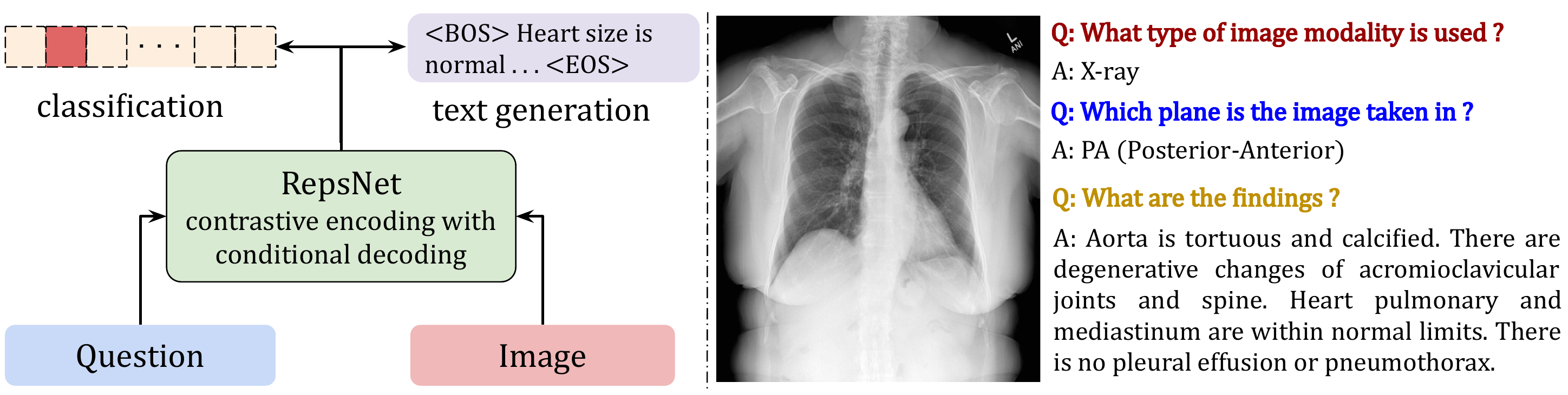}
\end{center}
\vspace{-0.6cm}
\caption{\footnotesize \textit{(left)} RepsNet analyzes medical images and automates report writing by providing answers to questions via classifying among known answer categories or generating natural language descriptions, \textit{(right)} radiology report generation example with top two categorical answers and bottom one natural language descriptive answer.}\label{fig: RepsNet_cover}
\vspace{-0.6cm}
\end{figure}
A long standing goal in artificial intelligence is to seamlessly interpret and describe medical images/videos with natural language. In this paper, we combine both vision and language modalities to interpret medical images in a visual question answering (VQA) setting, whereby we predict the answer to a given image and question using a novel encoder-decoder model (see Fig.~\ref{fig: RepsNet_cover}). We present RepsNet that fuses the encoded image and question features by contrastive alignment, while the decoder learns the conditional probability distribution to generate descriptions with: 1) encoded image and question features, and 2) prior context retrieved from nearest neighbouring reports of the image. We leverage publicly available ResNeXt~\cite{Xie_resnext_2016} and BERT~\cite{Devlin_bert_2018} for warm-starting the encoder, and GPT-2~\cite{Radford_GPT_2019} as the base model for the natural language decoder. 

We present its application to assist practitioners with automatic report generation from medical images~\cite{Jing_RepGen_2017,Chen_M2T_2020,Li_HybridRep_2018,Liu_cvpr_2021,Najdenkoska_MICCAI_2021}. Existing methods using hand-written notes, dictation services or electronic medical record templates are widely perceived to be time-consuming and cumbersome. To this end, we parse the medical report into a set of questions and handle both categorical (yes/no, multiple choices) and natural language descriptive answers (open-ended) in a visual question answering setting. We evaluate the proposed approach on two publicly available benchmark datasets: 1) visual question answering radiology (VQA-Rad) datasets in the span of $2018 - 2021$~\cite{Jason_vqa_rad_2018}, 2) Indiana University x-ray (IU-Xray) dataset containing chest x-ray images paired with reports describing findings and impressions~\cite{Demner_iuxray_2016}. RepsNet outperforms state-of-the-art models across both VQA-Rad and IU-Xray datasets.

\textbf{Contributions: }This paper makes three contributions:
\begin{itemize}
    \item We present RepsNet, an encoder-decoder model for writing reports that adapts pretrained models by contrastive alignment of images with answers in the encoding phase, and generates natural language descriptions by conditional decoding on images and prior context of retrieved reports.
    \item A visual question answering formulation to handle both categorical and natural language descriptive answers in generating automated reports. 
    \item Experiments on publicly available VQA-Rad and IU-Xray datasets with $81.08 \%$ classification accuracy and $0.58$ BLEU-$1$ score respectively, showing significant performance improvement over state-of-the-art methods.
\end{itemize}

\section{Related Work}
\textbf{Vision and Language Pretraining: }Self-supervised pre-training of language models such as BERT~\cite{Devlin_bert_2018}, GPT/GPT-2~\cite{Radford_GPT_2019}, XLNet have shown promising results in transferring knowledge across related tasks~\cite{Dong_nips_2019}. This has led to combining both visual and language modalities by cross-alignment of domains in a joint embedding space~\cite{Desai_VirTex_2020,Sariyildiz_icmlm_2020}. Examples include LXMERT~\cite{Tan_LXMERT_2019}, ViLBERT~\cite{Lu_VILBERT_2019}, PixelBERT~\cite{Zhicheng_pixelbert_2020}, VideoBERT~\cite{Sun_VideoBert_2019} and VisualGPT~\cite{Xia_xgpt_2020}. Authors in~\cite{Zhang_convirt_2020,Radford_clip_2021} use contrastive learning to pair images with textual descriptions as a whole, in contrast to grounding the masked works in the image locally in~\cite{Sun_CBT_2019,Gupta_arvix_2020}. To incorporate prior knowledge in pretrained language generation models~\cite{Yu_survey_2020}, Ziegler \textit{et al.}~\cite{Ziegler_condlang_2019} adapt a pretrained model for arbitrary source conditioning. Despite a few promising approaches, cross-domain conditioning of a pretrained model remains a challenge and can degrade the pretrained model representations. \newline
\textbf{Visual Question Answering and Image Captioning: }Describing medical images with visual question answering~\cite{Agarwal_vqa_15} or natural language~\cite{Xu_ShowTell_2015,Anderson_Caption_2017} is difficult due to rare and diverse nature of abnormalities, weak association of image features with text in reports, lack of prior domain knowledge, case-based reasoning, and long descriptions of findings. Medical VQA has recently received attention with small scale datasets such as VQA-Rad to categorize answers by classification~\cite{Nguyen_mevf_2019,Do_mmq_2021,Liu_CRPD_2021}. Several works have followed the image captioning line of work with an emphasis on generating long descriptions~\cite{Jing_RepGen_2017,Alfarghaly_cdgpt2_2021}, incorporating domain-specific medical knowledge~\cite{Zhang_aaai_2020,Liu_cvpr_2021}, retrieving description from a template~\cite{Li_HybridRep_2018,Liu_Rep_2019}, question answering~\cite{Ren_cgmvqa_2020,Sharma_MedFuse_2021}, among others~\cite{Jianbo_Miccai_2019,Najdenkoska_MICCAI_2021}. 

In this paper, we investigate the automated report writing under a novel visual question answering framework to handle both categorical and descriptive answers for a given image and a question. We use contrastive learning to align the paired images and report answers in an embedding space, and retrieve nearest neighbour report answers to incorporate prior knowledge in generating medical descriptions. 

\section{RepsNet: Proposed Approach}
\textbf{Problem Formulation: }
Given an image or a set of images $\mb{\mathrm{x}} \in \mb{\mathrm{X}}$, we are interested in generating a report comprising of $s$ answers $\mb{\mathrm{y}} = \{\mb{\mathrm{y}}_{1} \ldots \mb{\mathrm{y}}_{s}\} \in \mb{\mathrm{Y}}$, corresponding to the natural language questions $\mb{\mathrm{q}} = \{\mb{\mathrm{q}}_{1} \ldots \mb{\mathrm{q}}_{s}\} \in \mb{\mathrm{Q}}$. Each answer $\mb{\mathrm{y}}_i$ may be \textbf{close-ended} belonging to a fixed possible set of categories or \textbf{open-ended} comprising of multiple natural language sentences. Each word $\mb{\mathrm{w}} \in \mb{\mathrm{V}}$ in the open-ended answer belongs to a known natural language vocabulary. We seek to learn the model parameters $\mb{\mathrm{\Theta}}$ to maximize the conditional likelihood $\mathcal{P}_{\mb{\mathrm{\Theta}}}(\mb{\mathrm{y}}_i \; | \; \mb{\mathrm{x}}, \mb{\mathrm{q}}_i)$ of predicting the answers for a given image and a set of questions, 
\begin{equation}
\scalemath{0.9}{\mb{\mathrm{\Theta}} = \argmax_{\mb{\mathrm{\Theta}}} \sum_{i=1}^{s} \log \mathcal{P}_{\mb{\mathrm{\Theta}}}(\mb{\mathrm{y}}_i \; | \; \mb{\mathrm{x}}, \; \mb{\mathrm{q}}_i).}
\end{equation}
We formulate the problem with an encoder decoder model. The encoder  $\mb{f_{\theta_{\mathrm{enc}}}}: \{\mb{\mathrm{X, Q}}\} \rightarrow \{\mb{\mathrm{\bar{X}, \bar{Q}}}\} \in \mathbb{R}^{\{n_\mathrm{x}, n_\mathrm{q}\} \times \{d_\mathrm{x}, d_\mathrm{q}\}}$ transforms the image and the input text sequence to a joint cross-aligned visual and language representation space with $n_\mathrm{x}$ image pixels/regions, $n_\mathrm{q}$ text tokens, and $\{d_\mathrm{x}, d_\mathrm{q}\}$ hidden space dimensions of image and text embeddings respectively. The decoder $\mb{h_{\theta_{\mathrm{dec}}}}:\{\mb{\mathrm{\bar{X}, \bar{Q}, \bar{C}}}\} \rightarrow \mathcal{P}(\mb{\mathrm{Y}})$ models the conditional probability distribution of predicting the target answer $\mb{\mathrm{Y}}$ given the encoded hidden states $\{\mb{\mathrm{\bar{X}, \bar{Q}}}\}$, and the \textbf{prior context} $\mb{\mathrm{\bar{C}}} \in \mathbb{R}^{n_\mathrm{c} \times d_\mathrm{c}}$ of $n_\mathrm{c}$ tokens with dimension $d_\mathrm{c}$ that represents the domain specific knowledge for controlled text generation (we discuss prior context further in the next section). Note that we only use the prior context for generating open-ended answers.

\begin{figure*}[t]
\begin{center}
\includegraphics[trim={0.11cm 0.0cm 0.0cm 0.0cm},clip,scale=0.6]{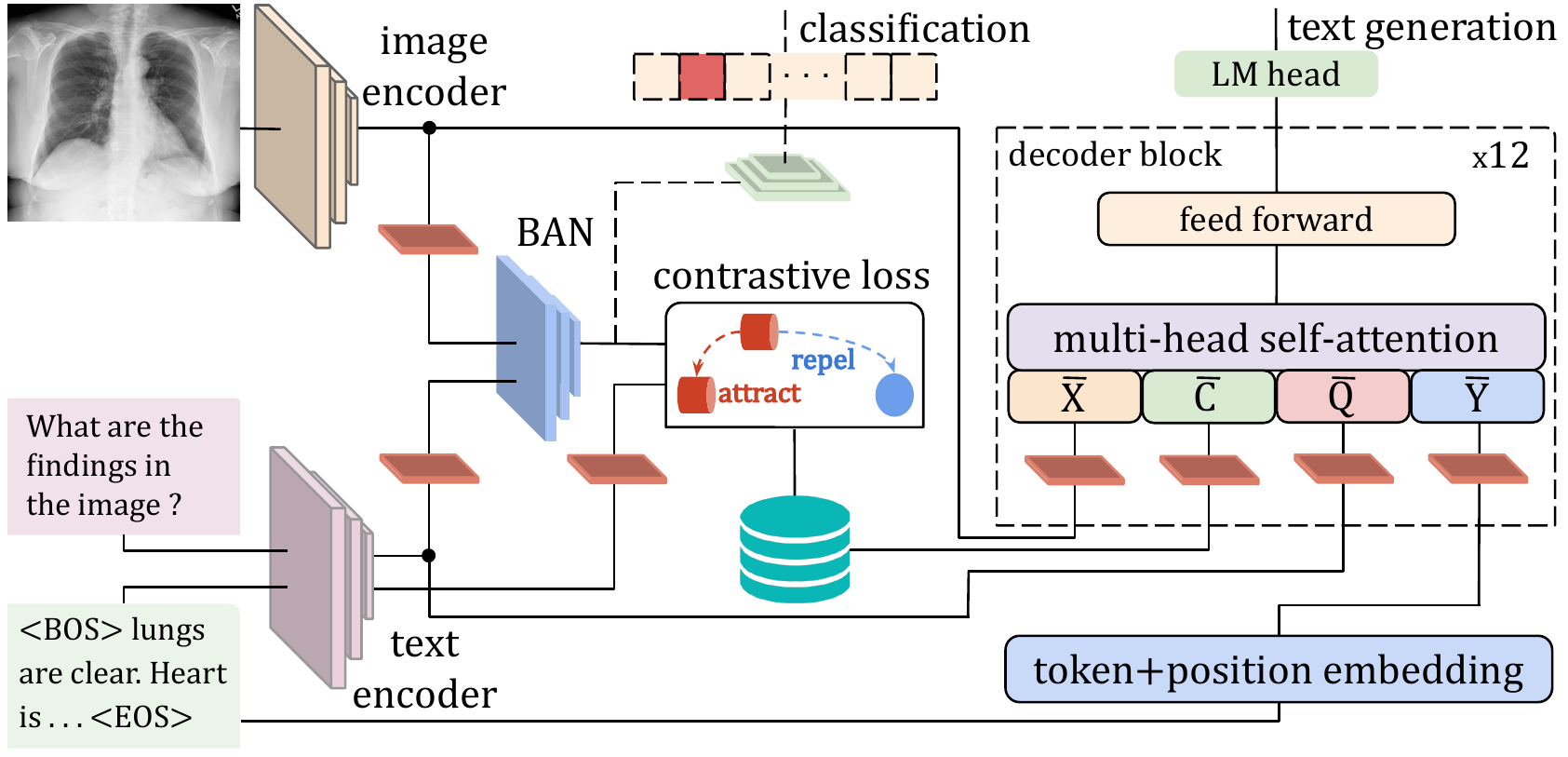}
\end{center}
\vspace{-0.6cm}
\caption{\footnotesize RepsNet encoded image and question features are fused via bilinear attention network (BAN), before self-supervised contrastive alignment with natural language descriptions. The answer is categorized via classification among fixed answer categories or generated by conditional language decoding on image, question and prior context of answers retrieved by nearest neighbour search. Note that we omit the question features $\mb{\mathrm{\bar{Q}}}$ in describing the conditional language decoder below for brevity.}\label{fig: RepsNet_architecture}
\vspace{-0.6cm}
\end{figure*}

In this paper, we leverage large-scale pretrained models for warm-starting the encoder and the decoder model parameters. For close-ended answers, we map the combined image and question features to the output layer of all possible close-ended answers for classification. For open-ended answers, the decoder retrieves the prior context $\mb{\mathrm{\bar{C}}}$ as the nearest neighbouring answers of the encoded image features, and greedily maximizes the learned conditional distribution $\mathcal{P}_{\mb{\theta_{\mathrm{dec}}}} \left(\mb{\mathrm{Y_t}} | \mb{\mathrm{Y_{0:t-1}, \bar{X}, \bar{Q}, \bar{C}}} \right)$ to generate the answer sequence $\mb{\mathrm{Y_{1:t}}}$ in an auto-regressive manner (see Fig.~\ref{fig: RepsNet_architecture}). 

\subsection{Contrastive Image-Text Encoder}
The encoder has four constituent parts: 1) \textbf{image encoder} to extract visual features, 2) \textbf{text encoder} to tokenize and contextualize natural language questions and answers features, 3) \textbf{bilinear attention network} to fuse the image and the question, and 4) \textbf{contrastive alignment} of visual features and textual answers. 

\textbf{Image Encoder: }We use the \texttt{ResNeXt-101}~\cite{Xie_resnext_2016} architecture as the base image encoder. We remove the last linear and pooling layer and add a 2D adaptive average pooling layer to resize the input image to a fixed feature space of $14 \times 14 \times 2048$ that preserves the correspondence between the visual features and the input image ($n_\mathrm{x}=196, d_\mathrm{x}=2048$). Moreover, we add image transformations, namely \textit{color jittering, normalization, random erasing}, to augment the training data distribution within each batch before extracting the visual features.

\textbf{Text Encoder: }We adapt the BERT~\cite{Devlin_bert_2018} model for the text encoder that is pre-trained to predict masked words locally based on the context provided by other non-masked words in the sequence. We filter out the punctuation marks and tokenize the text using WordPiece algorithm~\cite{Devlin_bert_2018}, before extracting the textual features.

\textbf{Bilinear Attention Network (BAN): }We use a BAN to fuse the cross-modal encoded question and image features~\cite{Kim_NIPS_2018}. The outer product or the bilinear product exhaustively combines the multi-modal features at the cost of higher computational complexity; in comparison to naive concatenation or inner product between the features. Compared to other co-attention mechanisms, BAN exploits bilinear interaction maps where each feature is pooled by low-rank bilinear approximations. Residual learning on top combines multiple bilinear attention maps for effective joint representation of question and image features.

For the sake of brevity and a slight abuse of notation, we denote $\mb{\mathrm{\bar{X}}}$ for both the image and the combined (image and question) features in describing the rest of the encoder and the decoder sections. 

\textbf{Contrastive Vision and Text Learning: }We align images with natural language descriptions via bidirectional contrastive learning~\cite{Chen_simclr_2020}, that pulls together a given image-answer pair, while pushing away observations that correspond to different image-answer pairs. 

Given the encoded image (and question) $\mb{\mathrm{\bar{X}}}$ and the natural language answer features $\mb{\mathrm{\bar{Y}}} \in \mathbb{R}^{n_\mathrm{y} \times d_{\mathrm{y}}}$ with $n_\mathrm{y}$ tokens of dimension $d_\mathrm{y}$, we first project them to a $d$-dimensional space with a linear transformation to $\mb{\mathrm{\hat{X}}} \in \mathbb{R}^{d}$ and $\mb{\mathrm{\hat{Y}}} \in \mathbb{R}^{d}$. During training, the loss operates on a mini-batch of $N_T$ image-text pairs $\{\mb{\mathrm{\hat{x}_i}}, \mb{\mathrm{\hat{y}_i}}\}_{i=1}^{N_T}$, where each pair is in turn taken as a positive sample to maximize agreement against all other negative samples, i.e.,
\begin{equation} \label{Eq: cos_sim}
    \scalemath{0.9}{\mathcal{L}_{\mb{\mathrm{\hat{x}}} \rightarrow \mb{\mathrm{\hat{y}}}} = - \frac{1}{N_T} \sum_{i=1}^{N_T} \; \log \; \frac{\exp \big(\langle \mb{\mathrm{\hat{x}_i}}, \mb{\mathrm{\hat{y}_i}} \rangle / \tau \big)}{ \sum_{j=1}^{N_T} \exp(\langle \mb{\mathrm{\hat{x}_i}}, \mb{\mathrm{\hat{y}_j}} \rangle / \tau) },}
\end{equation} where $\langle \mb{\mathrm{\hat{x}}}, \mb{\mathrm{\hat{y}}} \rangle = \frac{\mb{\mathrm{\hat{x}}}^{\trsp} \mb{\mathrm{\hat{y}}}}{ \Vert \mb{\mathrm{\hat{x}}} \Vert \Vert \mb{\mathrm{\hat{y}}}\Vert}$ represents the cosine similarity distance and $\tau \in \mathbb{R}^{+}$ represents the temperature parameter to scale the similarity metric. Similar to the image-to-text loss in Eq.~\eqref{Eq: cos_sim}, we also define the text-to-image loss $\mathcal{L}_{\mb{\mathrm{\hat{y}}} \rightarrow \mb{\mathrm{\hat{x}}}}$ to account for the asymmetry with respect to each input modality as in~\cite{Zhang_convirt_2020,Radford_clip_2021}. Overall bidirectional encoder loss $\mathcal{L}_{\mathrm{enc}}$ is the sum of the two constituent contrastive losses weighted by constant $\alpha_l \in \mathbb{R}^{+}$,
\begin{equation}
    \scalemath{0.9}{\mathcal{L}_{\mathrm{enc}} = \alpha_l (\mathcal{L}_{\mb{\mathrm{\hat{x}}} \rightarrow \mb{\mathrm{\hat{y}}}} +  \mathcal{L}_{\mb{\mathrm{\hat{y}}} \rightarrow \mb{\mathrm{\hat{x}}}}).}
\end{equation}

\textbf{Prior Context Knowledge: }We store the normalized natural language answers of the train set $\mb{\mathrm{\hat{Y}_{train}}}$ during model training. We then compute topk nearest neighbours $\mb{\mathrm{\bar{C}}}$ that maximize the cosine similarity between a given encoded image $\mb{\mathrm{\hat{X}}}$ and the stored natural language answers $\mb{\mathrm{\hat{Y}_{train}}}$. We use the \texttt{FAISS} library for scalable nearest neighbour search~\cite{Jeff_faiss_2017}. The prior context aids the decoder to attend to longer horizon dependencies and get additional case-based details for controlled text generation. This is particularly relevant in describing medical images with specific terminologies, writing style and class imbalanced abnormalities, i.e.,
\begin{equation}
    \scalemath{0.9}{\mb{\mathrm{\bar{C}}} \; = \; \mathrm{topk} \left[ \max_{i \in  \mb{\mathrm{\hat{Y}_{train}}}} \; \langle \; \mb{\mathrm{\hat{X}}} , \mb{\mathrm{\hat{Y}_{train}^{(i)}}} \rangle \right].}
\end{equation} 
\subsection{Conditional Language Decoder}

The probability distribution of generating the output text sequence $\mb{\mathrm{Y_{1:t}}}$ conditioned on the contextualized encoding sequence $\mathcal{P}_{\mb{\theta_{\mathrm{dec}}}} \left(\mb{\mathrm{Y_{1:t}}} | \mb{\mathrm{\bar{X}, \bar{C}}} \right)$ can be decomposed into a product of conditional distributions using Bayes' rule,
\begin{equation}
\scalemath{0.9}{\mathcal{P}_{\mb{\theta_{\mathrm{dec}}}} \left(\mb{\mathrm{Y_{1:t}}} | \mb{\mathrm{\bar{X}, \bar{C}}} \right) = \prod_{i=1}^{t} \mathcal{P}_{\mb{\theta_{\mathrm{dec}}}} \left( \mb{\mathrm{y_i}} | \mb{\mathrm{y_{0:i-1}}}, \mb{\mathrm{\bar{X}, \bar{C}}} \right),}\label{eq: cond_decode}
\end{equation} where $\mb{\mathrm{y_{0}}} = \langle \mb{\mathrm{BOS}} \rangle $ is a special token reserved for the beginning of a sentence. We model the conditional language generation with a stack of transformer-based blocks, using the GPT-2 model as the base pretrained language decoder~\cite{Radford_GPT_2019}. We introduce modifications to the GPT-2 model for conditioning on image and prior context features by directly adding their attention outputs to the pretrained self-attention layers of the model, similar to ~\cite{Ziegler_condlang_2019,Alfarghaly_cdgpt2_2021}, thereby adding the attention outputs for different conditional inputs with a parsimonious increase in the number of parameters only (see supplementary materials for details). The conditional probability distribution in Eq.~\eqref{eq: cond_decode} is maximized by optimizing the cross-entropy loss on the ground-truth and the predicted sequences.

\textbf{Overall Approach: }During training, we adapt the pretrained language and vision models in an end-to-end manner for contrastive encoding and conditional decoding with a small amount of image-text pairs. The overall training loss function comprises of the contrastive loss and the cross-entropy loss. During natural language generation, we predict the output sequence in an auto-regressive manner with greedy or beam search decoding, and stop generating the sequence once we predict a special end of text $\langle \mb{\mathrm{EOS}} \rangle $ token.

\section{Experiments, Results and Discussion}

We evaluate the performance of RepsNet in interpreting visual concepts on publicly available VQA-Rad~\cite{Jason_vqa_rad_2018} for classification and IU-Xray~\cite{Demner_iuxray_2016} for natural language generation. We are interested in evaluating: 1) how feasible it is to adapt the pretrained language and vision models in describing small set of medical images, 2) what is the role of contrastive encoding in learning joint visual linguistic representations, 3) does conditional decoding on image features and prior context help with generating medical language descriptions, and 4) how does RepsNet fare in performance among the state-of-the-art approaches.

\subsection{Visual Question Answering} 

\textbf{VQA-Rad Dataset: }We use the VQA-Rad datasets~\cite{Jason_vqa_rad_2018} from $2018-2019$ and also introduce an aggregrated dataset, VQA-Rad All, that combines all the VQA-Rad datasets from $2018-2021$. Radiology images in the datasets are taken from open-access MedPix database, and the questions are predominantly posed from categories, such as image plane, imaging modality, organ system involved and image abnormalities. The VQA problem is posed as a multi-class classification over all possible set of answers, and classification accuracy on evaluation set is used as the performance metric. We use the standard training and evaluation splits provided with the datasets (see summary in supplementary materials).\newline
\textbf{Results: }Table~\ref{tab: vqa_result} shows that RepsNet outperforms all other competing methods across all the datasets. Similar to other methods, RepsNet uses bilinear attention mechanism to fuse the image and the question features. Contrary to other methods, RepsNet does not use fixed Glove word embeddings~\cite{Jeffrey_glove_2014} or RNNs for sentence level representations; instead it learns the entire contextual embeddings using BERT-style transformer with WordPiece tokenization. Ablation study in Table~\ref{tab: ablation_vqa} also shows that the performance increases the most with the use of pre-trained models. We observe from Table~\ref{tab: vqa_result} that simply filtering out instances and class categories with less than $5$ and $10$ instances per class category $M_o = \{5, 10\}$ proportionally increases the classification accuracy across all datasets, at the cost of reducing the overall number of instances and class categories to mitigate class imbalance in the datasets. Note that we do not take into account unseen class category instances of the evaluation set in computing the classification accuracy.
\begin{table}[!tbp]
\parbox{.48\linewidth}{
\centering
\caption{\footnotesize Classification accuracy on the VQA-Rad datasets. Bottom three rows increase minimum occurrence threshold from $0$ to $5$ to $10$ instances. RepsNet outperforms all other competing methods.} \normalsize \centering \label{tab: vqa_result}
\vspace{-0.3cm}
\begin{tabular}{|c||c|c|c|}
\hline
 & \textbf{2018} & \textbf{2019}& \textbf{All} \\ \hline \hline
 \textbf{MEVF}~\cite{Nguyen_mevf_2019} & $66.10$  & $-$ & $-$ \\ \hline
\textbf{MMQ}~\cite{Do_mmq_2021} & $67.00$  & $-$ & $-$ \\ \hline
\textbf{QCR}~\cite{zhan_mvqa_2020} & $69.65$  & $-$ & $-$ \\ \hline 
\textbf{CLEF}~\cite{Abacha_ImageClef_2019} & $-$  & $62.40$ & $-$ \\ \hline 
\textbf{CRPD}~\cite{Liu_CRPD_2021} & $72.70$  & $-$ & $-$ \\ \hline \hline
\textbf{RepsNet}-$0$ & \cellcolor{gray!25} $\mb{81.08}$  & \cellcolor{gray!25} $\mb{67.57}$ & \cellcolor{gray!25} $\mb{63.69}$ \\ \hline 
\textbf{RepsNet}-$5$ & \cellcolor{gray!25} $\mb{83.55}$  & \cellcolor{gray!25} $\mb{79.83}$ & \cellcolor{gray!25} $\mb{71.93}$ \\ \hline
\textbf{RepsNet}-$10$ & \cellcolor{gray!25} $\mb{87.05}$  & \cellcolor{gray!25} $\mb{81.17}$  & \cellcolor{gray!25} $\mb{80.37}$ \\ \hline
\end{tabular}
\vspace{-0.3cm}
}
\hfill
\parbox{.5\linewidth}{
\centering
\vspace{-0.2cm}
\caption{\footnotesize BLEU scores (B1 - B4) for medical report generation on IU-Xray dataset. RepsNet yields better scores than other methods. } \normalsize \centering \label{tab: report_results}
\vspace{0.12cm}
\begin{tabular}{|c||c|c|c|c|}
\hline
 & \textbf{B1} & \textbf{B2} & \textbf{B3} & \textbf{B4} \\ \hline \hline
\textbf{Co-Att}~\cite{Jing_RepGen_2017} & $0.45$  & $0.29$ & $0.20$ & $0.15$ \\ \hline
\textbf{HRGR}~\cite{Li_HybridRep_2018} & $0.44$  & $0.30$ & $0.21$ & $0.15$ \\ \hline
\textbf{CMAS}~\cite{Jing_RepGen_2020} & $0.46$  & $0.30$ & $0.21$ & $0.15$ \\ \hline 
\textbf{Mem-T}~\cite{Chen_M2T_2020} & $0.47$  & $0.30$ & $0.22$ & $0.16$ \\ \hline
\textbf{VTI}~\cite{Najdenkoska_MICCAI_2021} & $0.49$  & $0.36$ & $0.29$ & $0.15$ \\ \hline 
\textbf{PPKED}~\cite{Liu_cvpr_2021} & $0.48$  & $0.31$ & $0.22$ & $0.17$ \\ \hline \hline
\textbf{RepsNet} & \cellcolor{gray!25} $\mb{0.58}$  & \cellcolor{gray!25} $\mb{0.44}$ & \cellcolor{gray!25} $\mb{0.32}$ & \cellcolor{gray!25} $\mb{0.27}$ \\ \hline 
\end{tabular}
}
\vspace{-0.3cm}
\end{table}

\subsection{Medical Report Generation}
\textbf{IU-XRay: }The Indiana University x-ray dataset~\cite{Demner_iuxray_2016} comprises of frontal and lateral views of chest x-ray images that are associated with radiology report sections, namely \textit{impressions, findings} and \textit{manual tags}. For brevity, we only report results for populating the findings question in this work, i.e., we associate the same question for all the answers. After omitting the reports without findings section, we randomly split the remaining $3607$ reports into $80\%$ training and $20\%$ evaluation sets. On average, each report instance has $5.7$ sentences, while each sentence has $6.5$ words that describe the image findings. Note that no classification labels are available to detect the anomalies. Max number of tokens for a report section is set to $200$ and the report findings are zero-padded in case its length is less than max number of tokens. We use the sentence level BLEU scores as the performance metric computed using the \texttt{nltk} library that compares $n$-gram similarity between the ground-truth and the generated report where $n$ varies from $1$ to $4$ (whereas for classification accuracy evaluation in VQA-Rad datasets, we compare the predicted and the ground-truth indices of the class categories).\newline
\begin{figure*}[t]
\begin{center}
\includegraphics[trim={0.0cm 0.0cm 0.0cm 0.0cm},clip,scale=0.55]{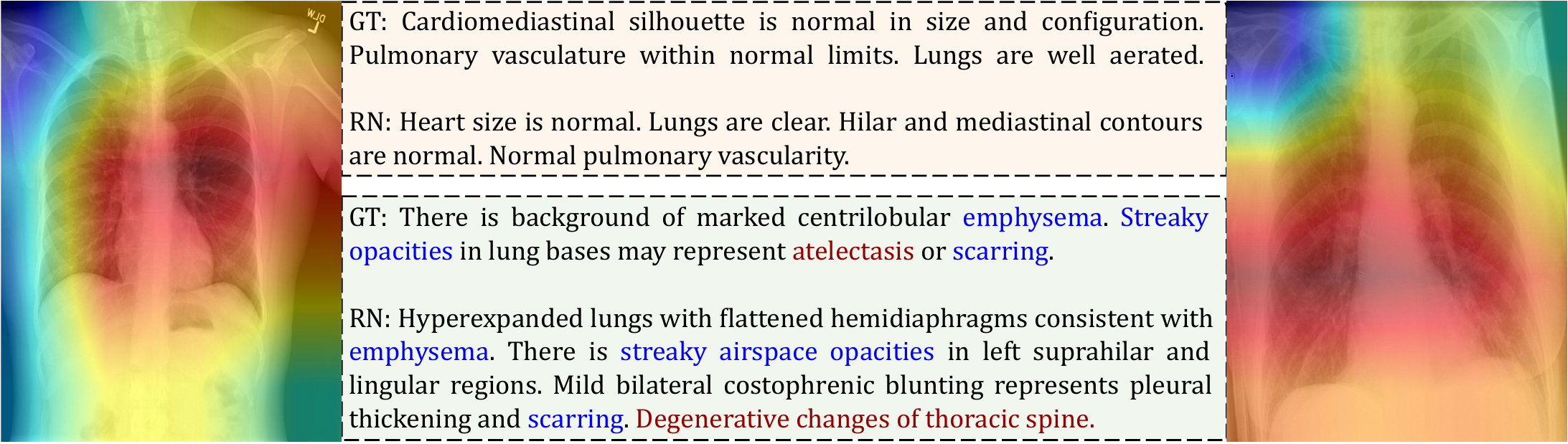}
\end{center}
\vspace{-0.6cm}
\caption{\footnotesize Heatmap visualization and comparison between ground-truth (GT) and RepsNet generated (RN) report of: \textit{(left)} normal case, \textit{(right)} abnormal case. RepsNet shows strong alignment with ground-truth in describing medical findings. Text in blue shows abnormalities, text in red represents misalignment.}\label{fig: RepsNet_qual}
\vspace{-0.2cm}
\end{figure*}
\begin{table}[tbp]
\parbox{.48\linewidth}{
\centering
\caption{\footnotesize Ablation study on VQA-Rad dataset to quantify the effect of pre-training, pre-processing and contrastive learning. Classification accuracy increases the most with pre-training, while pre-processing and contrastive learning stage further improve the performance.} \normalsize \centering \label{tab: ablation_vqa}
\vspace{-0.1cm}
\begin{tabular}{|c||c|c|c|c|}
\hline
\textbf{pretraining} & & \checkmark & \checkmark & \checkmark \\ \hline
\textbf{preprocess} &  &  & \checkmark & \checkmark \\ \hline
\textbf{contrastive} &  &  &  & \checkmark \\ \hline \hline
& $74.47$ & $79.12$ & $80.09$ & $\mb{81.08}$ \\\hline
\end{tabular}
\vspace{-0.5cm}
}
\hfill
\parbox{.5\linewidth}{
\centering
\caption{\footnotesize Ablation study on IU-Xray dataset with: \textit{(top)} visual features - Vis, \textit{(middle)} Vis with contrastive encoding - Vis + CE, and \textit{(bottom)} Vis with CE and prior context - Vis + CE + PC. BLEU scores improve with contrastive learning and prior context.} \normalsize \centering \label{tab: ablation_reports}
\vspace{-0.1cm}
\begin{tabular}{|c||c|c|c|c|}
\hline
 & \textbf{B1} & \textbf{B2} & \textbf{B3} & \textbf{B4} \\ \hline \hline
\textbf{Vis} & $0.48$  & $0.38$ & $0.30$ & $0.26$ \\ \hline
\textbf{Vis + CE} & $0.55$  & $0.42$ & $0.32$ & $0.27$ \\ \hline
\textbf{Vis + CE + PC} & $0.58$  & $0.44$ & $0.32$ & $0.27$ \\ \hline
\end{tabular}
\vspace{-0.5cm}
}
\end{table}
\textbf{Results: }Results are summarized in Table~\ref{tab: report_results}. It can be seen that RepsNet performs significantly better than the state-of-the-art report generation methods across all the BLEU scores, suggesting the feasibility of adapting large-scale pretrained language and vision models on a small set of domain-specific medical data. Ablation study in Table~\ref{tab: ablation_reports} reveals that adding visual features, contrastive learning and prior context subsequently boosts the performance of the GPT2-decoder. Fig.~\ref{fig: RepsNet_qual} provides a qualitative comparison between the ground-truth and the generated report findings, along with the heatmap visualizations using grad-cam~\cite{Selvaraju_GradCAM_2016} for an intuitive understanding of the approach. We observe a strong alignment in generating normal report findings, whereas part of the findings sometimes get omitted and/or added in describing the abnormalities, especially for rare cases (see supplementary materials for video demonstration and other examples). Systematic dealing of rare cases with external domain knowledge and past medical history of patients is a promising direction of our future work. We are also interested in incorporating attention mechanisms for conditional visualization of generated text on image patches as a measure of uncertainty in the prediction. Making these reports self-explainable is critical for its wider adoption. Other areas of interest include reducing the liability of generated report errors, as well as working with medical experts to evaluate the generated reports.
 
\section{Conclusion}

In this paper, we have presented RepsNet that adapts pre-trained vision and language models for describing a small set of domain-specific medical images. We take a unified visual question answering approach to predict class categories or generate descriptive answers for writing automated medical reports. RepsNet
is specifically tailored for contrastive alignment of images and text in the encoding phase, and combining visual and prior context of nearest neighboring reports with natural language generator in the decoding phase. This has enabled RepsNet to provide state-of-the-art results on challenging tasks of visual question answering and medical report generation on radiology images. In our future work, we plan to extend our approach to summarizing reports from videos, and transfer the developed methodology to clinical sites for automated reporting in gastroenterology.


%
%
%
\bibliographystyle{splncs04}
\bibliography{egbib}
\newpage
\appendix
\section{VQA-Rad Datasets}
\vspace{-0.5cm}
\begin{table*}[hbp]
\caption{\small Summary of VQA-Rad datasets from $2018-2021$: Table shows number of images (Im) and question-answer pairs (QA) in train and eval sets; number of classes $N_c$ and unseen instances in train $\mb{U_A^{(\mathrm{Train})}}$ and eval sets $\mb{U_A^{(\mathrm{Eval})}}$ as minimum occurrence $M_o$ of instances per class category increases from $0 \rightarrow 5 \rightarrow 10$. Class imbalance and unseen answers in the eval set make it difficult for VQA approaches.} \normalsize \centering \label{tab: vqa_rad_datasets}
\vspace{-0.2cm}
\begin{tabular}{|c||c|c||c|c||c|c|c||c|c|c||c|c|c|}
\hline
 & \multicolumn{2}{|c||}{\textbf{Train}} & \multicolumn{2}{|c||}{\textbf{Eval}} & \multicolumn{3}{|c||}{$\mb{N_c}$ w/ $\rightarrow \mb{M_o}$} & \multicolumn{3}{|c||}{$\mb{U_A^{\mathrm{Tr}}}$ w/ $\rightarrow \mb{M_o}$} & \multicolumn{3}{|c|}{$\mb{U_A^{\mathrm{Ev}}}$ w/ $\rightarrow \mb{M_o}$} \\ \cline{2-14} 
 & Im & QA & Im & QA & $0$ & $5$ & $ 10 $ & $0$ & $5$ & $ 10 $ & $0$ & $5$ & $ 10 $ \\ \hline \hline
\textbf{2018} & $315$  & $3064$ & $315$ & $451$& $458$ & $42$ & $10$ & $0$ & $936$ & $1194$ & $43$ & $141$ & $173$\\ \hline
\textbf{2019} & $3200$  & $12,792$ & $500$ & $2000$& $1540$ & $68$ & $36$ & $0$ & $3842$ & $4172$ & $150$ & $597$ & $651$\\ \hline
\textbf{2020} & $4000$  & $4000$ & $500$ & $500$& $331$ & $140$ & $2$  & $0$ & $1773$ & $3940$ & $0$& $230$ & $472$\\ \hline
\textbf{All} & $7437$  & $19,856$ & $1799$ & $3451$ & $2014$ & $270$ & $42$ & $0$ & $5245$ & $9300$ & $185$ & $975$ & $1795$ \\ \hline
\end{tabular}
\vspace{-0.3cm}
\end{table*}
\vspace{-0.5cm}
\section{Conditional Language Decoder Formulation}
Formally, the encoded input text sequence $\mb{\mathrm{\bar{Y}}}$ is linearly projected to the \textit{query}, \textit{key}, and \textit{value} vectors using respective projection matrices $\{\mb{\mathrm{W_{q\bar{y}}, W_{k\bar{y}}, W_{v\bar{y}}}}\} \in \mathbb{R}^{d_{\mathrm{y}} \times d_\mathrm{h}}$ of a decoder block. The conditioning encoder inputs $\mb{\mathrm{\bar{X}}}$ and $\mb{\mathrm{\bar{C}}}$ are then added to the key and the value vectors using pairs of projection matrices $\{\mb{\mathrm{W_{k\bar{x}}, W_{v\bar{x}}}}\} \in \mathbb{R}^{d_{\mathrm{x}} \times d_\mathrm{h}}$ and $\{\mb{\mathrm{W_{k\bar{c}}, W_{v\bar{c}}}}\} \in \mathbb{R}^{d_{\mathrm{c}} \times d_\mathrm{h}}$, respectively. The multi-modal self-attention matrix $\mathcal{A}(\mb{\mathrm{\bar{Y}, \bar{X}, \bar{C}}})$ for a decoder block can then be represented as a scaled dot-product,
\begin{equation}
    \mathcal{A} (\mb{\mathrm{\bar{Y}, \bar{X}, \bar{C}}}) = \texttt{softmax}\left( \left(\mb{\mathrm{\bar{Y}W_{q\bar{y}}}}\right) \begin{bmatrix}\mb{\mathrm{\bar{Y}W_{k\bar{y}}}} \\ \mb{\mathrm{\bar{X}W_{k\bar{x}}}} \\ \mb{\mathrm{\bar{C}W_{k\bar{c}}}}\end{bmatrix}^{\trsp}\right) \begin{bmatrix}\mb{\mathrm{\bar{Y}W_{v\bar{y}}}} \\ \mb{\mathrm{\bar{X}W_{v\bar{x}}}} \\ \mb{\mathrm{\bar{C}W_{v\bar{c}}}}\end{bmatrix}.
\end{equation} 

\section{Experimental Set-up}
\textbf{ VQA-Rad Experimental Setup: }We use the WordPiece tokenization method with a max token length of $12$ and pretrained BioBert -- BERT trained on PubMed articles -- to warm-start the text encoder. We use residual learning on top of bilinear attention networks using a glimpse of two projections, before joint alignment with the answer labels via contrastive learning. The decoder projects the encoded image and text sequence to a hidden dimension of $1024$ neurons before mapping it to classification categories of size equal to the number of answers in the dataset (see Table~\ref{tab: vqa_rad_datasets}). We use the standard train-eval splits provided with the datasets, Adam optimizer for fixed weight decay (AdamW) with a batch size of $64$ and a learning rate of $5\mathrm{e}{-5}$ for a total of $200$ epochs.

\textbf{ IU-Xray Experimental Setup: }We use the pretrained BERT and GPT-2 as base models for the encoder and the decoder respectively. BioBERT or ClinicalBERT did not improve report generation results in our experiments. Additional parameters for contrastive encoding and conditional decoding are randomly initialized. We use two separate optimizers for the encoder and the decoder parameters; each configured with the same AdamW optimizer with a batch size of $16$ and learning rate of $5\mathrm{e}{-5}$ that linearly decays over $100$ epochs. 

In the training phase, we learn the decoder parameters via \textbf{teacher forcing} where the target word is passed as the next input to the decoder and use cross-entropy loss to backpropagate the error between the ground-truth and the target sequences. During inference, we predict the next word via greedy search in a deterministic manner, while introducing penalties to ensure minimum length of the sequence is greater than $4$ and words are not repeated in the generation process. Moreover, we did not observe performance gains by sampling strategies such as top-k and/or top-k with top-p nucleus sampling. We use the ground-truth report as prior context during training, and include $1$ nearest neighbour report as prior context during evaluation. For more details, see the qualitative analysis of generated reports below and the deployment results in the supplementary video.

\section{IU-Xray Report Generation Examples}

\vspace{-0.5cm}
\begin{figure*}[hbp]
\begin{center}
\vspace{-0.3cm}
\includegraphics[trim={0.0cm 0.0cm 0.0cm 0.0cm},clip,scale=0.58]{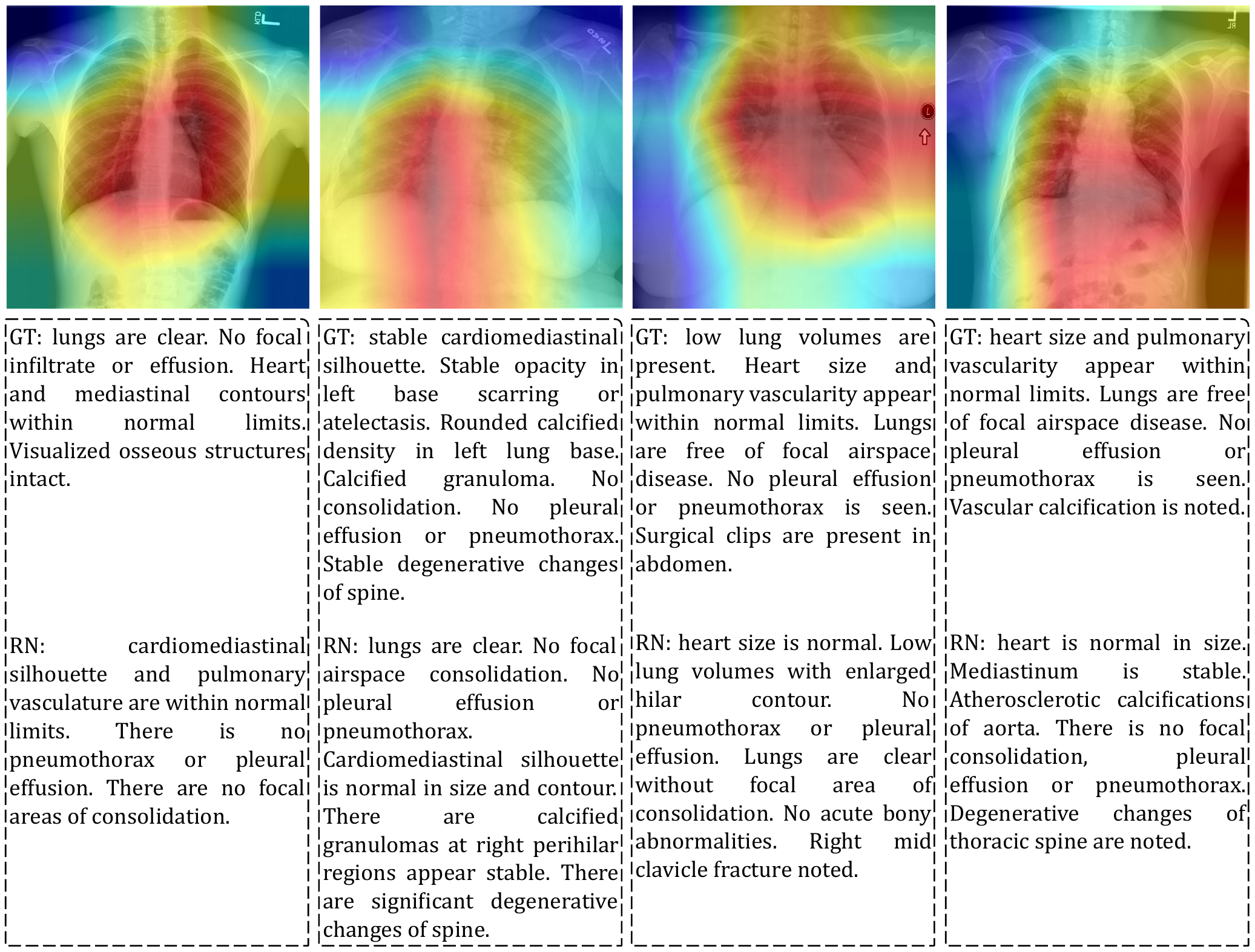}
\end{center}
\vspace{-0.3cm}
\caption{\footnotesize Heatmap visualization and comparison between ground-truth (GT) and RepsNet generated (RN) report. RepsNet shows strong alignment with ground-truth in describing medical findings.}\label{fig: RepsNet_qual}
\end{figure*}

\end{document}